\documentclass[%
reprint,
superscriptaddress,
nofootinbib,
amsmath,amssymb,
aps,
]{revtex4-2}

\usepackage{soul}
\usepackage{graphicx}
\usepackage{dcolumn}
\usepackage{bm}
\usepackage{hyperref}
\hypersetup{
	colorlinks=true,
	citecolor=blue,
	linkcolor=blue,
	urlcolor=blue,
}

\DeclareMathOperator*{\argmin}{argmin}

\newcommand{\rIto}{\:\tilde{\cdot}\:}

\begin{document}


\title{Sch\"odinger Bridge Type Diffusion Models\\ as an Extension of Variational Autoencoders}

\author{Kentaro Kaba}
    \affiliation{%
    Department of Physics, Institute of Science Tokyo, Meguro-ku, Tokyo 152-9551, Japan
    }%
\author{Reo Shimizu}%
    \affiliation{%
    Graduate School of Information Sciences, Tohoku University, Sendai, Miyagi 980-9564, Japan
    }%
\author{Masayuki Ohzeki}%
    \affiliation{%
    Department of Physics, Institute of Science Tokyo, Meguro-ku, Tokyo 152-9551, Japan
    }%
    \affiliation{%
    Graduate School of Information Sciences, Tohoku University, Sendai, Miyagi 980-9564, Japan
    }%
    \affiliation{%
    Sigma-i Co., Ltd., Minato-ku, Tokyo 108-0075, Japan
    }%
\author{Yuki Sughiyama}%
    \affiliation{%
    Graduate School of Information Sciences, Tohoku University, Sendai, Miyagi 980-9564, Japan
    }%

\date{\today}

\begin{abstract}
    Generative diffusion models use time-forward and backward stochastic differential equations to connect the data and prior distributions. While conventional diffusion models (e.g., score-based models) only learn the backward process, more flexible frameworks have been proposed to also learn the forward process by employing the Schr\"odinger bridge (SB). However, due to the complexity of the mathematical structure behind SB-type models, we can not easily give an intuitive understanding of their objective function. In this work, we propose a unified framework to construct diffusion models by reinterpreting the SB-type models as an extension of variational autoencoders. In this context, the data processing inequality plays a crucial role. As a result, we find that the objective function consists of the prior loss and drift matching parts.
\end{abstract}

\maketitle

\paragraph*{Introduction.---\hspace{-9pt}}\label{sec:introduction}
Generative diffusion model~\cite{sohl2015deep,ho2020denoising,song2020score,wang2021deep,vargas2021solving,de2021diffusion,chen2022likelihood,tong2024improving} is one of the likelihood-based generative models that have successfully synthesized high-quality samples such as image~\cite{dhariwal2021diffusion,batzolis2021conditional} and audio~\cite{chen2020wavegrad,popov2021grad}. 
In the framework of the diffusion model, the model distribution that mimics the training dataset is defined by a distribution transported from a simple prior through a stochastic differential equation (SDE). 
This framework enables us to generate new samples by learning the transport dynamics via neural networks (NNs).

A prominent example of generative diffusion models is the score-based model (SBM)~\cite{song2020score,song2021maximum}. In SBMs, the data distribution is gradually perturbed by the addition of noise and eventually converges to a normal distribution that does not have any information. 
This procedure is commonly referred to as the noising process. Subsequently, the reversal of the noising process is learned to establish a generative process, which synthesizes new samples by transporting the prior distribution (i.e., the normal distribution).
Although SBMs are used in many situations~\cite{dhariwal2021diffusion,batzolis2021conditional, chen2020wavegrad,popov2021grad, cao2024survey}, they have two main limitations. 
First, since learning of the reverse process requires computation of the score function (the gradient of the log probability distribution), 
the noising process must be restricted to a simple form (e.g., linear or degenerate), 
where the score function can be evaluated analytically. 
Second, when the data distribution far deviates from the prior, the time required for the noising process becomes longer,
which results in substantial delays in sample generation.

To overcome these limitations, a more flexible model has recently been proposed using the Schrödinger bridge (SB)~\cite{wang2021deep, vargas2021solving, de2021diffusion,chen2022likelihood, tong2024improving}. The SB is a stochastic optimization technique that derives an SDE to connect any two arbitrary probability distributions over a given time horizon~\cite{schrodinger1931uber,caluya2021wasserstein}. 
This technique allows us to use highly structured SDEs with NNs for the noising process and acceletates the generation of new samples. 
However, the mathematical structure underlying SB-type diffusion models is complicated, because it involves mapping partial differential equations that solve SB problems to SDEs of noising and generative processes by using the non-linear Feynman-Kac formula~\cite{chen2022likelihood,exarchos2018stochastic}. 
This complicated structure inhibits our intuitive understanding of the objective function used in the training of SB-type models.

In this work, we attempt to reinterpret SB-type models as a relatively straightforward extension of variational autoencoders (VAEs)~\cite{kingma2013auto,kingma2019intro}. 
VAEs are generative models that learn two mappings between data and latent spaces: the encoder that maps the data into the latent variables and the decoder that represents its inverse. 
Recent work~\cite{luo2022understanding} has rederived SBMs in the VAE framework by associating the noising process with the encoder and the generative process with the decoder, respectively. However, the approach in that work has omitted the training of the encoder and has limited the scope to the derivation of SBMs. 
In this letter, taking into account the training of the encoder, we propose a method to construct the objective function for SB-type models.
As a result, it is revealed that the objective function is composed of the prior loss and drift matching parts, which characterize the training of NNs in the encoder and the decoder, respectively.


\paragraph*{Variational Autoencoders.---\hspace{-9pt}}\label{sec:VAE}
We begin with a brief introduction to variational autoencoders.
Let $\{x_i\}_{i=1}^n \in \mathcal{X}^n$ be a dataset
and define its empirical distribution in $\mathcal{X}$ as
$\mu(x) = (1/n)\sum_i \delta(x-x_i)$.
$q_\theta(x)$ denotes a generative model with a parameter $\theta$.
In addition, the training of the model based on the data distribution $\mu(x)$ is performed by solving
\begin{align}
        \theta^* 
        &= \argmin_\theta D_\text{KL}(\mu(x)\|q_\theta(x)),
        \label{eq:objective_function}
\end{align}
which represents the minimization of the Kullback-Leibler (KL) divergence: $D_\text{KL}(\mu\|q_\theta) := \mathbb{E}_{\mu}[\log (\mu/q_\theta)]$. 
That is equivalent to the maximum log-likelihood training~\cite{mitchell1997machine}.

In VAEs, the model $q_\theta(x)$ is defined by the marginal distribution of $Q_\theta(x,z)$ as
\begin{align}
    q_\theta(x) := \int dz\,Q_\theta(x,z) = \int dz\,Q_\theta(x|z)\pi(z).
    \label{eq:q_theta_VAE}
\end{align}
Here, $\pi(z)$ is a known distribution of a latent variable $z \in \mathcal{Z}$, 
which is called the {prior}\footnote{
    We often choose that $\dim \mathcal{Z} \leq \dim \mathcal{X}$ and the prior is a normal distribution.
};
$Q_\theta(x|z)$ is a conditional distribution of $x$ given $z$. 
Since this distribution carries information from the latent space $\mathcal{Z}$ to the data space $\mathcal{X}$, it is called the decoder.

In VAEs,  instead of solving Eq.~\eqref{eq:objective_function} directly, we consider the optimization problem of its upper bound by employing the data processing inequality\footnote{This inequality usually describes the relationships among mutual informations, but in this paper, we apply it for the KL divergences.} (Appendix~\ref{sec:appendix_DP_ineq}):
\begin{align}
    D_\text{KL}(\mu(x)\|q_\theta(x)) \leq D_\text{KL}(P_\phi(x, z)\|Q_\theta(x,z)).
    \label{eq:data_processing_inequality}
\end{align}
That minimizes the KL divergence on the joint space $\mathcal{X} \times \mathcal{Z}$. 
Here, $P_\phi(x, z)$ is the joint distribution extended from $\mu(x)$ with a conditional distribution $P_\phi(z|x)$:
\begin{align}
    P_\phi(x, z) = P_\phi(z|x)\mu(x).
    \label{eq:P_phi}
\end{align}
In contrast to the decoder $Q_\theta(x|z)$, since $P_\phi(z|x)$ sends the information from $\mathcal{X}$ to $\mathcal{Z}$ with the parameter $\phi$, it is called the {encoder}.
The training of VAEs is to find the optimal parameters $(\phi^*, \theta^*)$: 
\begin{align}
    (\phi^*, \theta^*) = \argmin_{\phi, \theta} D_\text{KL}(P_\phi(x, z)\|Q_\theta(x,z)).
    \label{eq:objective_function_VAE}
\end{align}
By taking Eqs.~\eqref{eq:q_theta_VAE} and~\eqref{eq:P_phi} into account, the objective function of VAEs is derived as 
\begin{align}
    \begin{split}
        &D_\text{KL}(P_\phi(x,z)\|Q_\theta(x,z))
        =\mathbb{E}_{\mu(x)}[\log \mu(x)]\\
        &+\mathbb{E}_{\mu(x)}[D_\text{KL}({P}_\phi(z|x)\|\pi(z))]
        -\mathbb{E}_{\mu(x){P}_\phi(z|x)}\!\left[\log Q_\theta(x|z)\right]\!.
        \label{eq:decomposition_VAE}
    \end{split}
\end{align}
Its derivation is shown in Appendix~\ref{sec:appendix_ELBO_VAE}.
The negative sign of the second and third terms is collectively referred to as the evidence lower bound~\cite{kingma2019intro}.
In practical training, the encoder $P_\phi(z|x)$ and the decoder $Q_\theta(x|z)$ are usually
modeled by a normal distribution whose mean and variance are given by neural networks (NNs).

After training, we can generate new samples using the optimized parameter $\theta^*$ as follows:
(i) The latent variable $z$ is sampled from a known prior $\pi(z)$.
(ii) A new sample $x$ can be generated by the optimal decoder $Q_{\theta^*}(x|z)$ conditioned on $z$.
Here, we note that $\phi^*$ is not used in the generating process but is used for the training of $\theta$.

To clarify the scheme of this training, we summarize the role of each parameter.
On the one hand, training of $\theta$ adjusts the model distribution $q_\theta$ closer to the data distribution $\mu$, which is the primary goal of all generative models.
On the other hand, training for $\phi$ does not directly contribute to optimizing $q_\theta$ because $\phi$ does not appear in the LHS of Eq.~\eqref{eq:data_processing_inequality}.
That serves more to fill the gap between both sides of the inequality and helps to improve the accuracy of the training of $\theta$.
This structure of training will play an important role in revealing the mechanism of diffusion models.

\paragraph*{Diffusion models in the framework of VAEs.---\hspace{-9pt}}\label{sec:diffusion_model}
In this work, we regard the generative diffusion models as an extension of VAEs.
The key idea behind this extension is based on the fact that the data processing inequality~\eqref{eq:data_processing_inequality} 
holds regardless of the number of latent variables in the joint distribution.
Therefore, the number of variables can be extended to infinity, which forms a path along which the probability can be analyzed.

We start with the setup of diffusion models. 
Let $x_{[0,T]} = \{x_t\}_{t\in[0,T]}$ be a path in the time interval $[0,T]$, and 
$\mathbb{Q}_\theta[x_{[0,T)}|x_T=z]$ denotes a conditional path probability given the final state $x_T = z$.
Since $\mathbb{Q}_\theta[x_{[0,T)}|x_T=z]$ carries the information on the latent space $\mathcal{Z}$\footnote{
In diffusion models, $\dim \mathcal{Z} = \dim \mathcal{X}$
} along the path $x_{[0,T]}$ time-backwardly, it works as the {decoder}.
Setting a prior to $\pi(z)$, the probability of the path $x_{[0,T]}$ is expressed as $\mathbb{Q}_\theta[x_{[0,T]}]=\mathbb{Q}_\theta[x_{[0,T)}|x_T=z]\pi(z)$.
Then, the diffusion model $q_\theta(x)$ is defined by the marginal distribution at the initial state $x_0$:
\begin{align}
    q_\theta(x) 
    := \int \prod_{t\in[0,T]} dx_t\, \mathbb{Q}_\theta[x_{[0,T]}]\delta(x_0 - x).
    \label{eq:q_theta_DM}
\end{align}

Similarly to the previous section, to employ the data processing inequality, we define an encoder on the path $x_{[0,T]}$;
$\mathbb{P}_\phi[x_{(0,T]}|x_0 = x]$ denotes a conditional path probability for the encoder. 
Note that $\mathbb{P}_\phi[x_{(0,T]}|x_0 = x]$ is conditioned on the initial state $x_0$ of the path, 
whereas the decoder $\mathbb{Q}_\theta[x_{[0,T)}|x_T=z]$ is conditioned on the final state $x_T$.
Using the data distribution $\mu(x)$, we obtain the path probability $\mathbb{P}_\phi[x_{[0,T]}] = \mathbb{P}_\phi[x_{(0,T]}|x_0 = x]\mu(x)$.

In the diffusion models, the encoder $\mathbb{P}_\phi[x_{(0,T]}|x_0 = x]$ and the decoder $\mathbb{Q}_\theta[x_{[0,T)}|x_T=z]$ are modeled by the following time-forward and backward stochastic differential equations (SDEs) respectively:
\begin{gather}
    dX_t = u_\phi(t,X_t)\cdot d{t} + g(t)\cdot d{w}_t, 
    \label{eq:SDE_encoder}\\
    dX_t = s_\theta(t, X_t) \rIto dt + g(t)\rIto d\bar{w}_t, 
    \label{eq:SDE_decoder}
\end{gather}
where $u_\phi(t, X_t)$ and $s_\theta(t, X_t)$ represent the drifts of the diffusion processes.
They are given by certain NNs whose parameters are $\phi$ and $\theta$.
$g(t)$ denotes a given noise intensity and $w_t$ and $\bar{w}_t$ are independent Wiener processes~\cite{gardiner2009stochastic,risken1996fokker}.
The two types of products in SDEs~\eqref{eq:SDE_encoder} and~\eqref{eq:SDE_decoder} represent the It\^o and the reverse-It\^o products~\cite{gardiner2009stochastic,hirono2024a}, which are defined as
\begin{align}
    f(t, X_t)\cdot dw_t &:= f(t, X_{t})(w_{t+dt} - w_t),\\
    f(t, X_t)\rIto dw_t &:= f(t+dt, X_{t+dt})(w_{t+dt} - w_t).
\end{align}
The decode SDE~\eqref{eq:SDE_decoder} should be expressed in the reverse-It\^o products because, 
in our setting, the decoder is conditioned on $X_T=z$ and described in the time-backward manner. 
For the numerical generation of new samples, we need to calculate a current event from a future event, which is only feasible with the reverse-It\^o product.

The data processing inequality~\eqref{eq:data_processing_inequality} also can be applied to the case of path probabilities (see Appendix~\ref{sec:appendix_DP_ineq}):
\begin{align}
    D_\text{KL}(\mu(x)\|q_\theta(x)) \leq D_\text{KL}(\mathbb{P}_\phi[x_{[0,T]}]\|\mathbb{Q}_\theta[x_{[0,T]}]).
    \label{eq:data_processing_inequality_DM}
\end{align}
Hence, if we use the same framework in VAEs, the objective function of the diffusion models is given by the KL divergence between the path probabilities. That is, we solve the following optimization problem:
\begin{align}
    (\phi^*,\theta^*) = \argmin_{\phi,\theta} D_\text{KL}(\mathbb{P}_\phi[x_{[0,T]}]\|\mathbb{Q}_\theta[x_{[0,T]}]).
    \label{eq:objective_function_DM}
\end{align}

For the practical training of the diffusion models, we represent $D_\text{KL}(\mathbb{P}_\phi[x_{[0,T]}]\|\mathbb{Q}_\theta[x_{[0,T]}])$ by the NNs: $u_\phi$ and $s_\theta$.
To this end, we need to align the time direction of the encoder $\mathbb{P}_\phi[x_{(0,T]}|x_0 = x]$ and the decoder $\mathbb{Q}_\theta[x_{[0,T)}|x_T=z]$.
To reverse the time direction of the encoder, we define the conditional distribution given that $x_T =z$ by Bayes' theorem:
\begin{align}
    \mathbb{P}_\phi[x_{[0,T)}|x_T = z] = \frac{\mathbb{P}_\phi[x_{[0,T]}]}{p_\phi(z)},
    \label{eq:Bayes}
\end{align}
where $p_\phi(z)$ is the marginal distribution at the final state $x_T$ (i.e., the latent variable $z$):
\begin{align}
    p_\phi(z) 
    := &\int \prod_{t\in[0,T]} dx_t\, \mathbb{P}_\phi[x_{[0,T]}]\delta(x_T - z).
    \label{eq:p_phi_DM}
\end{align}
The conditional probability $\mathbb{P}_\phi[x_{[0,T)}|x_T=z]$ can be evaluated
by the time-reverse-encode SDE corresponding to Eq.~\eqref{eq:SDE_encoder}
with the initial sample $X_T\sim p_\phi$:
\begin{align}
    dX_t = [u_\phi(t, X_t) - g(t)^2\nabla \log \rho_\phi(t,X_t)]\rIto dt + g(t)\rIto d\bar{w}_t. 
    \label{eq:SDE_encoder_reverse}
\end{align}
Here, $\rho_\phi(t,\cdot)$ is 
the solution of the Fokker-Planck equation (FPE)~\cite{gardiner2009stochastic,risken1996fokker} with initial condition $\rho(0,\cdot) = \mu(\cdot)$:
\begin{gather}
    \frac{\partial \rho(t,x)}{\partial t}  
    = -\nabla \cdot [u_\phi(t, x)\rho(t,x)]
    + \frac{1}{2}g(t)^2 \nabla^2 \rho(t,x),
    \label{eq:FPE_forward}
\end{gather}
which expresses the time evolution of the probability distribution corresponding to the encode SDE~\eqref{eq:SDE_encoder}.
The derivation of Eqs.~\eqref{eq:SDE_encoder_reverse} and \eqref{eq:FPE_forward} is shown in Appendix~\ref{sec:appendix_reverse_OMfunc}.
In the context of diffusion models, $\nabla \log \rho_\phi(t,X_t)$ is known as a score function.

By using Eq.~\eqref{eq:Bayes}, the objective function in Eq.~\eqref{eq:objective_function_DM} is calculated as 
\begin{align}
    \begin{split}
        &D_\text{KL}(\mathbb{P}_\phi[x_{[0,T]}]\|\mathbb{Q}_\theta[x_{[0,T]}])
        = D_\text{KL}(p_\phi(z)\|\pi(z))\\
        &+ \mathbb{E}_{p_\phi(z)}\left[D_\text{KL}(\mathbb{P}_\phi[x_{[0,T)}|x_T=z]\|\mathbb{Q}_\theta[x_{[0,T)}|x_T=z])\right].
        \label{eq:objective_function_DM_decomposed}
    \end{split}
\end{align}
Moreover, by using the Girsanov theorem \cite{oksendal2013stochastic} for the reverse-It\^o process, the second term can be reformulated as
\begin{align}
    &\mathbb{E}_{p_\phi(z)}\left[ D_\text{KL}(\mathbb{P}_\phi[x_{[0,T)}|x_T=z]\|\mathbb{Q}_\theta[x_{[0,T)}|x_T=z])\right]\nonumber\\
    &=\frac{1}{2}\int_0^T \frac{dt}{g(t)^2}\mathbb{E}_{\rho_\phi(t,x_t)}\big[\|u_\phi(t, x_t)\nonumber\\
    &\qquad\qquad\qquad-g(t)^2\nabla \log \rho_\phi(t,x_t)-s_\theta(t, x_t)\|^2\big],
    \label{eq:Girsanov}
\end{align}
(see Appendix~\ref{sec:appendix_objective_DM}).
Therefore, our training scheme~\eqref{eq:objective_function_DM} is rephrased as
\begin{widetext}
    \begin{align}
        (\phi^*, \theta^*)
        = \argmin_{\phi, \theta} \left\{D_\text{KL}(p_\phi(z)\|\pi(z)) + 
        \frac{1}{2}\int_0^T \frac{dt}{g(t)^2}\mathbb{E}_{\rho_\phi(t,x_t)}\!\left[\|u_\phi(t, x_t)-g(t)^2\nabla \log \rho_\phi(t,x_t)-s_\theta(t, x_t)\|^2\right] \right\},
        \label{eq:objective_function_DM_reformed}
    \end{align}
\end{widetext}
which is the main result of this work. 
By solving Eq.~\eqref{eq:objective_function_DM_reformed}, we obtain the optimal parameters $(\phi^*, \theta^*)$ in NNs. 
Similarly to VAE, we can generate new samples by using the parameter $\theta^*$ in the decoder:
(i) Sample the latent variable $z$ from a known prior $\pi(z)$.
(ii) By time-backwardly solving the decode SDE~\eqref{eq:SDE_decoder} with $\theta^*$ under the terminal condition $X_T=z$, we obtain a new sample $x$ as $X_0 = x$.
For the same reason as in VAEs, the parameter $\phi^*$ in the encoder is not directly related to the generative processes,
but the training of $\phi$ makes the upper bound in Eq.~\eqref{eq:data_processing_inequality_DM} approach $D_\text{KL}(\mu\|q_\theta)$.

\begin{figure}[htbp]
    \includegraphics[width=\linewidth]{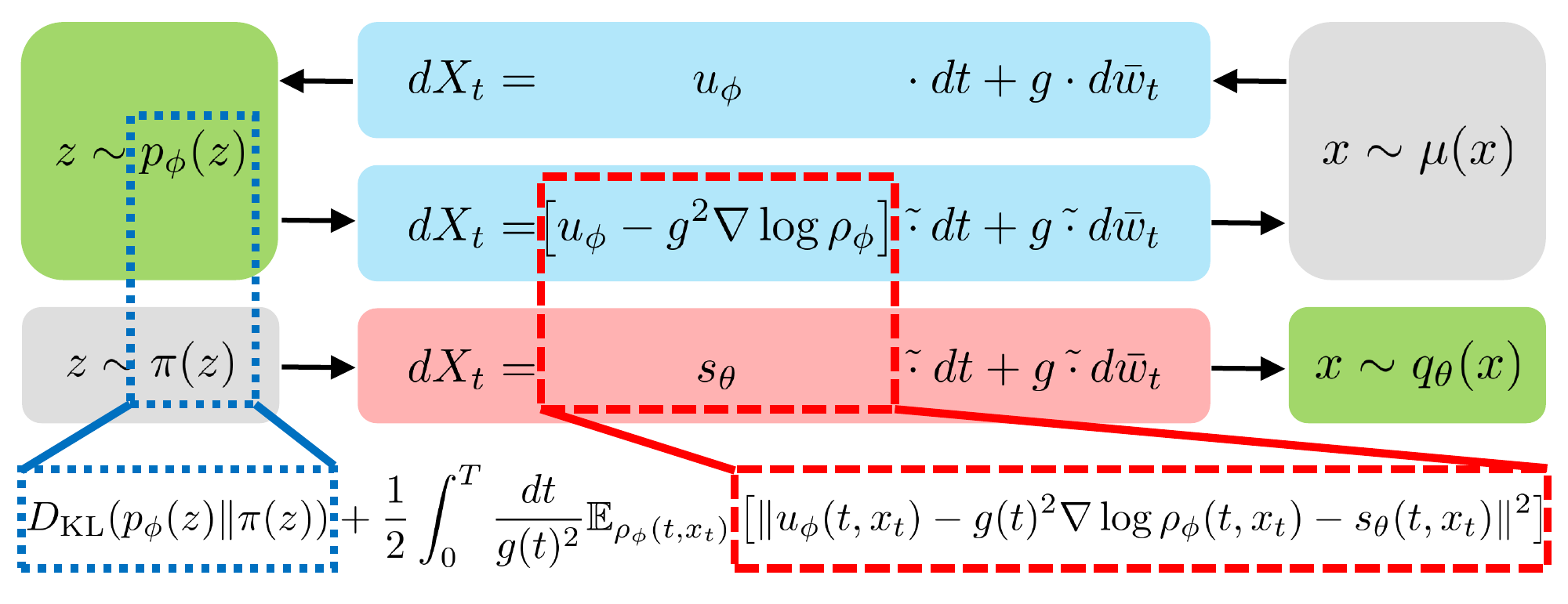}
    \caption{
        The top and middle SDEs with the blue shadow denote the encode SDE with NN $u_\phi$ and its time reversal, which connect the data $\mu$ and $p_\phi$. 
        By contrast, the bottom SDE with the red shadow represents the decode SDE with the other NN $s_\theta$, which transports the prior $\pi$ to $q_\theta$. 
        The first term in Eq.~\eqref{eq:objective_function_DM_reformed} means the prior loss between $p_\phi$ and $\pi$ (see blue dotted squares); the second term indicates the drift matching between the reverse-encode and the decode SDEs (see red dotted squares).
        }
    \label{fig:objective}
\end{figure}

This framework inspired from VAEs lays the foundation for a construction of diffusion models.
As shown in the next section, we can derive the training scheme of SBMs and SB-type models in this framework.
The previous attempt to interpret diffusion models as an extension of VAEs~\cite{luo2022understanding} has been limited to the model that does not learn the encoder and has derived only SBMs.
By contrast, our framework imposes no such restriction, which enables us to obtain the training scheme of SB-type models.

The objective function in Eq.~\eqref{eq:objective_function_DM_reformed} also gives us the interpretation for the training of diffusion models as follows (see Fig.~\ref{fig:objective}).
The first term can be considered as the prior loss. 
Training the parameter $\phi$ in the encoder implies searching an SDE which transports the data distribution $\mu(x)$ to the prior $\pi(z)$ by minimizing the KL divergence.
The second term can be interpreted as the drift matching.
If we find the optimal parameter $\phi^*$, we can generate new samples by solving the reverse-encode SDE~\eqref{eq:SDE_encoder_reverse} with $\phi^*$.
Nevertheless, the numerical generation of new samples is difficult since the drift term of the reverse-encode SDE includes the score function $\nabla \log \rho_\phi$.
The second term exists for mimicking the drift term of the reverse-encode SDE with the other NN in the decoder, $s_\theta(t,X_t)$.

\paragraph*{Connection to previous work.---\hspace{-9pt}}\label{sec:othermodels}
We can rederive the previous works with new perspectives by employing the framework established in the above section. 
In this letter, we deal with two models and one sampling method.

\textbf{Score-based models.}
In our framework, score-based models~\cite{song2020score, song2021maximum} can be understood as models that omit the training of the parameter $\phi$ in the encoder. 
To skip the training of $\phi$, we need to manually find an encode SDE such that the prior loss in Eq.~\eqref{eq:objective_function_DM_reformed} vanishes.

Consider an encode SDE (i.e., $u_\phi = f$ in Eq.~\eqref{eq:SDE_encoder}): 
\begin{align}
    dX_t = f(t, X_t)\cdot dt + g(t)\cdot dw_t,
    \label{eq:SDE_encoder_f}
\end{align}
where the drift $f(t, \cdot)$ is tuned as having the global stationary distribution $p_\text{st}$; i.e., for any initial distribution, the corresponding probability distribution $\rho_t$ satisfies $\lim_{t\to\infty} \rho_t = p_\text{st}$.
Then, by setting the prior $\pi = p_\text{st}$ and taking the time interval $T$ in the diffusion model as the limit $T\to \infty$, we obtain $\lim_{T\to \infty} D_\text{KL}(\rho_T\|\pi)=0$. Therefore, if such $f$ and $g$ are adopted, the objective function in Eq.~\eqref{eq:objective_function_DM_reformed} is reduced to 
\begin{align}
    \begin{split}
        \frac{1}{2}\int_0^\infty dt\, \frac{1}{g(t)^2}&\mathbb{E}_{\rho(t,x_t)}\left[\|f(t,x_t)\right.\\
        &\left.-g(t)^2\nabla \log \rho(t,x_t) - s_\theta(t, x_t)\|^2\right],
    \end{split}
    \label{eq:objective_function_DM_SBM}
\end{align}
where the expectation is taken with respect to the encode SDE Eq.~\eqref{eq:SDE_encoder_f}. Furthermore, by defining a new decode NN $s'_\theta := (s_\theta + f)/g^2$, Eq.~\eqref{eq:objective_function_DM_SBM} can be rewritten as
\begin{align}
    \frac{1}{2}\int_0^\infty dt\,g(t)^2\mathbb{E}_{\rho(t,x_t)}\!\big[\|s'_\theta(t,x_t) - \nabla\log\rho(t,x_t)\|^2\big].
    \label{eq:objective_function_SBM_reformed}
\end{align}
This objective function coincides with that of SBMs, known as \textit{explicit} score matching~\cite{hyvarinen2005estimation}.
In usual SBMs, the encode SDE~\eqref{eq:SDE_encoder_f} is modeled by a simple form (e.g., linear or degenerate drift), and the prior $\pi$ is chosen as a normal distribution.
Since this combination vanishes the prior loss in the limit $T\to \infty$, this usual setting allows us to use Eq.~\eqref{eq:objective_function_DM_SBM} for the training. 

In practical training of SBMs, we face two problems: the computation of the score function $\nabla \log \rho$ and the infinite-time-horizon optimization of Eq.~\eqref{eq:objective_function_SBM_reformed}.
The former is settled by using the technique called the \textit{denoising} score matching~\cite{vincent2011connection} if the encode SDE~\eqref{eq:SDE_encoder_f} can be analytically solved.
The latter is addressed by truncating the integral range in Eq.~\eqref{eq:objective_function_SBM_reformed} to a finite time interval that can be feasible in a numerical computation.
This means that the prior loss does not exactly vanish during practical training.
Therefore, we can conclude that the training of SBMs is an approximation to reduce computational costs at the expense of accuracy of Eq.~\eqref{eq:objective_function_DM_reformed}.

\textbf{SB-FBSDE models.}
To realize the practical training without the above approximation, diffusion models based on the Schrödinger bridge have been recently developed~\cite{wang2021deep, vargas2021solving, de2021diffusion,chen2022likelihood, tong2024improving}.
SB-type models achieve high learning accuracy in a finite-time horizon by incorporating an NN into the noising (encoding) process Eq.~\eqref{eq:SDE_encoder_f}.
The SB-FBSDE model~\cite{chen2022likelihood} is one of the  SB-type models and is rigorously established.
Its objective function is derived by using the nonlinear Feynman-Kac theorem~\cite{exarchos2018stochastic}, which tells us the SDEs corresponding to the partial differential equations representing the optimal condition of the SB problem~\cite{caluya2021wasserstein}.
This model can also be understood within our framework as follows.

It is possible to rewrite the objective function in Eq.~\eqref{eq:objective_function_DM_reformed}, by the technique in the \textit{implicit} score matching~\cite{hyvarinen2005estimation}, as 
\begin{widetext}
    \begin{align}
            (\phi^*, \theta^*)
            = \argmin_{\phi, \theta} \bigg\{-\mathbb{E}_{p_\phi(z)}[\log \pi(z)] 
            + \frac{1}{2}\int_0^T \frac{dt}{g(t)^2}\mathbb{E}_{\rho_\phi(t,x_t)}\left[\|u_\phi(t, x_t)-s_\theta(t, x_t)\|^2_2
            -2g(t)^2\nabla\cdot s_\theta(t,x_t)\right] \bigg\}.
        \label{eq:objective_function_DM_computation}
    \end{align}
\end{widetext}
Its derivation is shown in Appendix~\ref{sec:appendix_reformed_objective}.
Similarly to the SBM case, if we replace the drift terms as $(u_\phi, s_\theta) = (f + g^2u_\phi', f - g^2s_\theta')$ by defining new encode and decode NNs: $u_\phi'$ and $s'_\theta$, Eq.~\eqref{eq:objective_function_DM_computation} agrees with the optimization in the SB-FBSDE model (see theorem 4 in the reference~\cite{chen2022likelihood}). This implies that the learning process of the SB-FBSDE model is composed of the minimization of the prior loss and the drift matching. 
Conversely, this result also shows that our proposed training scheme Eq.~\eqref{eq:objective_function_DM_reformed} implicitly solves the SB problem.
Moreover, since Eq.~\eqref{eq:objective_function_DM_computation} does not contain the score function $\nabla\log\rho_\phi$ explicitly, we can numerically solve it in the finite horizon $T$. 

\textbf{Probability-flow ODE.}
The probability-flow ODE \cite{song2020score, song2021maximum,chen2022likelihood} is a sampling method proposed for diffusion models, which enables us to generate new samples with ordinary differential equation (ODE) instead of SDE.
This method is rederived in our framework as follows.

By using the FPE corresponding to the decode SDE~\eqref{eq:SDE_decoder}, an ODE equivalents to the SDE~\eqref{eq:SDE_decoder} can be derived as
\begin{align}
    \frac{dX_t}{dt} = s_\theta(t, X_t)+ \frac{1}{2}g(t)^2\nabla\log\tilde{\rho}_\theta(t,X_t),
    \label{eq:probability-flow_ODE_1}
\end{align}
where $\tilde{\rho}_\theta$ is the marginal distribution of $\mathbb{Q}_\theta[x_{[0,T]}]$ at time $t$ (see Appendix~\ref{sec:appendix_probability-flow_ODE}).
Under the optimal condition of Eq.~\eqref{eq:objective_function_DM_reformed}, the score function is approximated by the difference of the drift terms: $g^2\nabla \log \tilde{\rho}_{\theta^*} \simeq u_{\phi^*} - s_{\theta^*}$ (see the second term). 
Then, Eq.~\eqref{eq:probability-flow_ODE_1} can be rewritten as 
\begin{align}
    \frac{dX_t}{dt} \simeq \frac{1}{2}[u_{\phi^*}(t, X_t)+s_{\theta^*}(t, X_t)].
    \label{eq:probability-flow_ODE}
\end{align}
This is the probability-flow ODE for SB-type models. 
Unlike generation via the SDE~\eqref{eq:SDE_decoder}, generation via the ODE requires two parameters $(\phi^*, \theta^*)$.
If we manually set $u_\phi = f$ and shift the NN as $s_\theta = f- g^2s_\theta'$ as in SBMs,
Eq.~\eqref{eq:probability-flow_ODE} reduces to
\begin{align}
    \frac{dX_t}{dt} \simeq f(t, X_t) - \frac{1}{2}g(t)^2s'_{\theta^*}(t, X_t),
\end{align}
which is the probability-flow ODE for SBMs.

As the ODE can be solved deterministically, it enables faster sampling than the SDE. 
However, Eq.~\eqref{eq:probability-flow_ODE} is an approximation as long as the objective function in Eq.~\eqref{eq:objective_function_DM_reformed} does not vanish.
Therefore, the probability-flow ODE is more susceptible to the effects of the training loss, and the generative process is expected to be less stable.

\paragraph*{Conclusion.---\hspace{-9pt}}\label{sec:conclusion}
We have proposed an alternative method for deriving the objective function of SB-type diffusion models as the extension of VAEs. Our objective function~\eqref{eq:objective_function_DM_reformed} demonstrates that the diffusion model training scheme is structured by minimizing the prior loss and performing the drift matching. The key insight lies in extending the number of latent variables from one to infinity, leveraging the data processing inequality. This perspective enables us to interpret SB-type models within the framework of VAEs.

In this letter, we have modeled the encoder $\mathbb{P}_\phi[x_{[0,T]}|x_0]$ and the decoder $\mathbb{Q}_\theta[x_{[0,T]}|x_T]$ by SDEs~\eqref{eq:SDE_encoder} and \eqref{eq:SDE_decoder}, but this is just one example.
Even if we extend them to other probabilistic rules (e,g., non-Markov processes),
the proposed derivation for the objective function is expected to be applicable.  
A consequent training scheme of the objective function may be composed of two parts:
One learns the encoding process that transports the data to the prior, corresponding to the prior loss, and the other trains the decoding process to mimic the reversal of the encoding process, which is the drift matching.
\\
\paragraph*{Acknowledgments.---\hspace{-9pt}}
This study was supported by the JSPS KAKENHI Grant No. 23H01432.
Our study received financial support from the public\verb|\|private R\&D investment strategic expansion prograM (PRISM) and programs for bridging the gap between R\&D and ideal society (Society 5.0) and generating economic and social value (BRIDGE) from the Cabinet Office.

\appendix
\section{Data Processing Inequality}\label{sec:appendix_DP_ineq}
We derive the data processing inequality.
Consider the two joint probability distributions $P(\{x_i\}_{i=1}^N)$, $Q(\{x_i\}_{i=1}^N)$ and the KL divergence between them:
\begin{multline}
    D_\text{KL}(P(\{x_i\}_{i=1}^N)\|Q(\{x_i\}_{i=1}^N)) \\= \mathbb{E}_{P(\{x_i\}_{i=1}^N)}\left[\log \frac{P(\{x_i\}_{i=1}^N)}{Q(\{x_i\}_{i=1}^N)}\right].
    \label{eq:A1}
\end{multline}
The joint distributions can be written as the product of conditional and marginal distributions: 
\begin{align}
    P(\{x_i\}_{i=1}^N) = P(x_N|\{x_i\}_{i=1}^{N-1})P(\{x_i\}_{i=1}^{N-1}),
    \label{eq:A2}\\
    Q(\{x_i\}_{i=1}^N) = Q(x_N|\{x_i\}_{i=1}^{N-1})Q(\{x_i\}_{i=1}^{N-1}).
    \label{eq:A3}
\end{align}
The substitution of Eqs.~\eqref{eq:A2} and~\eqref{eq:A3} into Eq.~\eqref{eq:A1} leads to 
\begin{align}
    &D_\text{KL}(P(\{x_i\}_{i=1}^N)\|Q(\{x_i\}_{i=1}^N))\nonumber\\
    &= D_\text{KL}(P(\{x_i\}_{i=1}^{N-1})\|Q(\{x_i\}_{i=1}^{N-1})) \nonumber\\
    &+ \mathbb{E}_{P(\{x_i\}_{i=1}^{N-1})}[D_\text{KL}
        (P(x_N|\{x_i\}_{i=1}^{N-1})\|Q(x_N|\{x_i\}_{i=1}^{N-1}))
    ].
\end{align}
Since the second term of RHS is non-negative, we obtain the data processing inequality:
\begin{multline}
    D_\text{KL}(P(\{x_i\}_{i=1}^N)\|Q(\{x_i\}_{i=1}^N))\\
    \geq D_\text{KL}(P(\{x_i\}_{i=1}^{N-1})\|Q(\{x_i\}_{i=1}^{N-1})).
\end{multline}
If we set $N=2$, this reduces to Eq.~\eqref{eq:data_processing_inequality}. Furthermore, by repeating the above procedure, we reach
\begin{align}
    D_\text{KL}(P(\{x_i\}_{i=1}^N)\|Q(\{x_i\}_{i=1}^N)) \geq D_\text{KL}(P(x_1)\|Q(x_1)), 
\end{align}
which corresponds to Eq.~\eqref{eq:data_processing_inequality_DM} in the limit $N \to \infty$.
\section{The objective function for VAEs}\label{sec:appendix_ELBO_VAE}
The Equation~\eqref{eq:decomposition_VAE} is derived as follows.
By using the definitions of the encoder Eq.~\eqref{eq:P_phi} and the decoder Eq.~\eqref{eq:q_theta_VAE}, we can calculate the KL divergence as
\begin{align}
    &D_\text{KL}(P_\phi(x,z)\|Q_\theta(x,z))
    =\mathbb{E}_{P_\phi(x,z)}\left[ \log\frac{P_\phi(x,z)}{Q_\theta(x,z)} \right]\nonumber\\
    &= \mathbb{E}_{P_\phi(z|x)\mu(x)}\left[\log\frac{P_\phi(z|x)\mu(x)}{Q_\theta(x|z)\pi(z)} \right]\nonumber\\
    &= \mathbb{E}_{P_\phi(z|x)\mu(x)}\left[\log\mu(x)+ \log\frac{P_\phi(z|x)}{\pi(z)}-\log Q_\theta(x|z)\right].
    \label{eq:appendix_KL_divergence_joint_reformed}
\end{align}
Hence, the objective function can be decomposed as
\begin{align}
    \begin{split}
        &D_\text{KL}(P_\phi(x,z)\|Q_\theta(x,z))
        =\mathbb{E}_{\mu(x)}[\log \mu(x)]\\
        &+\mathbb{E}_{\mu(x)}[D_\text{KL}({P}_\phi(z|x)\|\pi(z))]
        -\mathbb{E}_{\mu(x){P}_\phi(z|x)}\!\left[\log Q_\theta(x|z)\right]\!.
    \end{split}
    \label{eq:appendix_decomposition_VAE}
\end{align}

\section{Derivation of Onsager-Machlup action and time-reverse SDE}\label{sec:appendix_reverse_OMfunc}

Before deriving Eqs.~\eqref{eq:SDE_encoder_reverse} and~\eqref{eq:FPE_forward},
we show how to calculate the conditional path probabilities from the It\^o and reverse-It\^o SDEs.

First, we consider the It\^o SDE:
\begin{align}
    dX_t = f(t,X_t)\cdot dt + g(t)\cdot d{w}_t,
    \label{eq:appendix_C_forward_SDE}
\end{align}
where $w_t$ is the standard Wiener process.
For an infinitesimal time interval $[t, t + \Delta t]$, Eq.~\eqref{eq:appendix_C_forward_SDE} can be discretized as
\begin{align}
    X_{t+\Delta t} - X_{t} &= f(t ,X_{t})\Delta t + g(t)\Delta{w}_{t}.
\end{align}
Here, $\Delta {w}_t = w_{t+\Delta t} - w_{t}$, and it is sampled from the normal distribution:
\begin{align}
    \mathcal{N}(\Delta w_t; 0, \Delta t) = \frac{1}{\sqrt{2\pi \Delta t}}\exp\left(-\frac{\|\Delta w_t\|^2}{2\Delta t}\right).
\end{align}
Therefore, the transition probability from $X_t = x_t$ to $X_{t + \Delta t} = x_{t + \Delta t}$ is computed as
\begin{align}
    &T(x_{t+\Delta t}|x_t) = \frac{\Delta x_{t+\Delta t}}{\sqrt{2\pi g(t)^2\Delta t}}\nonumber\\
    &\times \exp\left[-\frac{\Delta t}{2g(t)^2}\left\|\frac{x_{t+\Delta t} - x_t}{\Delta t} - f(t,x_t)\right\|^2\right].
    \label{eq:appendix_C_transition_probability}
\end{align}
Also, the conditional path probability $\mathbb{P}[x_{(0,T]}|x_0]$ is written as
\begin{align}
    &\mathbb{P}[x_{(0,T]}|x_0] \prod_{t\in[0,T]} dx_t
    = \lim_{N\rightarrow \infty} \prod_{i=0}^{N-1}T(x_{i+1}|x_i)\nonumber\\
    &= \lim_{N\rightarrow \infty} \prod_{i=0}^{N-1}\frac{\Delta x_{i+1}}{\sqrt{2\pi g(t_i)^2\Delta t}}\nonumber\\
    &\times \exp \left[ - \Delta t \sum_{i=0}^{N-1} \frac{1}{2g(t_i)^2}\left\|\frac{x_{i+1} - x_i}{\Delta t} - f(t_i,x_i)\right\|^2\right].
    \label{eq:appendix_C_path_probability}
\end{align}
Here,
$x_i = x_{t_i}, t_i = i\Delta t$ and $\Delta t = T/N$.
Then, by formally writing 
$\dot{x}_tdt := x_{t+dt}-x_t, \mathcal{D}x := \prod_{t \in [0,T]} dx_t/\sqrt{2\pi g(t)^2 dt}$, 
we can express Eq.~\eqref{eq:appendix_C_path_probability} as the path integral formulation:
\begin{gather}
    \mathbb{P}[x_{(0,T]}|x_0] \prod_{t\in[0,T]} dx_t = \mathcal{D}x\, e^{-\int_0^T dt \mathcal{L}(t, x_t, \dot{x}_t)},\label{eq:encoder_pathprobform}\\
    \mathcal{L}(t,x_t, \dot{x}_t) = \frac{1}{2g(t)^2}\left\|\dot{x}_t - f(t,x_t)\right\|^2.
\end{gather}
The function $\mathcal{L}(t, x_t, \dot{x}_t)$ is known as the Onsager-Machlup Lagrangian \cite{risken1996fokker,hirono2024a}.

Second, we culculate the conditional path probability $\tilde{\mathbb{P}}[x_{[0,T)}|x_T]$ generated from the reverse-It\^o SDE:
\begin{align}
    dX_t = f(t,X_t)\rIto dt + g(t)\rIto d{w}_t.
    \label{eq:appendix_C_backward_SDE}
\end{align}
This can be discretized as
\begin{align}
    X_{t+\Delta t} - X_{t} &= {f}(t+\Delta t ,X_{t+\Delta t})\Delta t + {g}(t+\Delta t)\Delta{w}_{t}.
\end{align}
Corresponding to Eq.~\eqref{eq:appendix_C_transition_probability}, the transition probability $\tilde{T}(x_t|x_{t+\Delta t})$ from $X_{t + \Delta t} = x_{t + \Delta t}$ to $X_t = x_t$ is computed as
\begin{align}
    &\tilde{T}(x_t|x_{t+\Delta t}) = \frac{\Delta x_{t}}{\sqrt{2\pi g(t+\Delta t)^2\Delta t}}\nonumber\\
    &\times \exp\!\left[-\frac{\Delta t}{2g(t+\Delta t)^2}\left\|\frac{x_{t+\Delta t} - x_t}{\Delta t} - f(t+\Delta t,x_{t+\Delta t})\right\|^2\right]\!.
\end{align}
Hence, the conditional path probability $\tilde{\mathbb{P}}[x_{[0,T)}|x_T]$ is written as
\begin{align}
    &\tilde{\mathbb{P}}[x_{[0,T)}|x_T] \prod_{t\in[0,T]} dx_t = \lim_{N\rightarrow \infty} \prod_{i=0}^{N-1}\tilde{T}(x_i|x_{i+1})\nonumber\\
    &= \lim_{N\rightarrow \infty} \prod_{i=0}^{N-1}\frac{\Delta x_{i}}{\sqrt{2\pi g(t_{i+1})^2\Delta t}}\nonumber\\
    &\times \exp \left[ -  \sum_{i=0}^{N-1} \frac{\Delta t}{2g(t_{i+1})^2}\left\|\frac{x_{i+1} - x_i}{\Delta t} - f(t_{i+1},x_{i+1})\right\|^2\right].
    \label{eq:appendix_C_path_probability_reverse_descritized}
\end{align}
By using a Taylor expansion for $f$ and $g$ with respect to time $t$ and ignoring the terms with irrelevant order of $\Delta t$, Eq.~\eqref{eq:appendix_C_path_probability_reverse_descritized} is further annlyzed as follows.
The infinitesimal transition probability $\tilde{T}(x_t|x_{t+\Delta t})$ is calculated as
\begin{align}
    &\tilde{T}(x_t|x_{t+\Delta t})\nonumber\\
    &\approx \frac{\Delta x_{t}}{\sqrt{2\pi g(t)^2\Delta t}}\nonumber\\
    &\quad\times \exp \Bigg[ -\frac{\Delta t}{2g(t)^2}\bigg\|\frac{x_{t+\Delta t} - x_t}{\Delta t} - f(t,x_{t}) - \partial_t f(t, x_t)\Delta t\nonumber\\
    &\qquad  - \nabla \cdot f(t, x_t)(x_{t+\Delta t} - x_t)^2 - \frac{1}{2}g(t)^2\nabla^2f(t, x_t)\Delta t\bigg\|^2 \Bigg]\nonumber\\
    &= \frac{\Delta x_{t}}{\sqrt{2\pi g(t)^2\Delta t}}\nonumber\\
    &\times \exp \left[ -\frac{\Delta t}{2g(t)^2}\left\|\frac{x_{t+\Delta t} - x_t}{\Delta t} - f(t,x_{t})\right\|^2 + \nabla\cdot f(t, x_t)\Delta t \right].
    \label{eq:appendix_C_transition_probability_reverse}
\end{align}
Here, we used the It\^o formula~\cite{gardiner2009stochastic} for the expansion of $f$ and $g$. 
One may be concerned that the It\^o formula can not be applied to this Taylor expansion since $x_{t+\Delta t}$ is not a stochastic variable. 
However, a realization $x_{[0,T]}$ which does not satisfy the SDE~\eqref{eq:appendix_C_backward_SDE} never occurs. Thus, we can formally use the It\^o formula.

By pluging Eq.~\eqref{eq:appendix_C_transition_probability_reverse} into Eq.~\eqref{eq:appendix_C_path_probability_reverse_descritized},  the conditional path probability $\tilde{\mathbb{P}}[x_{[0,T)}|x_T]$ is also written as
\begin{align}
    &\tilde{\mathbb{P}}[x_{[0,T)}|x_T] \prod_{t\in[0,T]} dx_t = \lim_{N\rightarrow \infty} \prod_{i=0}^{N-1}\tilde{T}(x_i|x_{i+1})\nonumber\\
    &= \lim_{N\rightarrow \infty} \prod_{i=0}^{N-1}\frac{\Delta x_{i}}{\sqrt{2\pi g(t_i)^2\Delta t}}\exp \Bigg[ - \Delta t\nonumber\\
    &\times  \sum_{i=0}^{N-1}\left\{  \frac{1}{2g(t_i)^2}\left\|\frac{x_{i+1} - x_i}{\Delta t} - f(t_i,x_i)\right\|^2- \nabla\cdot f(t_i ,x_i) \right\}\Bigg].
\end{align}
In the path integral formulation, we find 
\begin{gather}
    \tilde{P}[x_{[0,T)}|x_T] \prod_{t\in[0,T]} dx_t = \mathcal{D}x\, e^{-\int_0^T dt \tilde{\mathcal{L}}(t, x_t, \dot{x}_t)},
    \label{eq:appendix_C_path_integral_reverse}\\
    \tilde{\mathcal{L}}(t,x_t, \dot{x}_t) = \frac{1}{2g(t)^2}\left\|\dot{x}_t - f(t,x_t)\right\|^2 - \nabla\cdot f(t, x_t).
\end{gather}
By using the above two path probabilities, Eqs.~\eqref{eq:encoder_pathprobform} and~\eqref{eq:appendix_C_path_integral_reverse}, we derive Eqs.~\eqref{eq:SDE_encoder_reverse} and~\eqref{eq:FPE_forward} as follows.
The numerator in the RHS of Eq.~\eqref{eq:Bayes}, $\mathbb{P}_\phi[x_{[0,T]}]$, can be represented by the following path integral formulation:
\begin{align}
    \mathbb{P}_\phi[x_{[0,T]}]\prod_{t\in[0,T]} dx_t 
    &= \mathcal{D}x\,\mu(x_0) e^{-\int_0^T dt \mathcal{L}_{\phi}(t,x_t, \dot{x}_t)},
    \label{eq:appendix_D_path_integral}
\end{align}
where $\mathcal{L}_\phi(t,x_t, \dot{x}_t)$ is the Onsager-Machlup Lagrangian:
\begin{align}
    \mathcal{L}_\phi(t,x_t, \dot{x}_t) = \frac{1}{2g(t)^2}\left\|\dot{x}_t - u_\phi(t,x_t)\right\|^2.
\end{align}
By substituting Eq.~\eqref{eq:appendix_D_path_integral} into Eq.~\eqref{eq:Bayes}, we have
\begin{align}
    \mathbb{P}_\phi[x_{[0,T)}|x_T]\prod_{t\in[0,T]} dx_t
    &= \mathcal{D}x\,\frac{\mu(x_0)}{p_\phi(x_T)} e^{-\int_0^T dt \mathcal{L}_{\phi}(t,x_t, \dot{x}_t)}\nonumber\\
    &= \mathcal{D}x\, e^{-\int_0^T dt \mathcal{L}'_{\phi}(t,x_t, \dot{x}_t)},
\end{align}
where $\mathcal{L}'_{\phi}(t,x_t, \dot{x}_t)$ is defined as
\begin{align}
    \mathcal{L}'_{\phi}(t,x_t, \dot{x}_t) = \mathcal{L}_{\phi}(t,x_t, \dot{x}_t) + \frac{d}{dt}\log \rho_\phi(t, x_t).
\end{align}
Furthermore,  the second term can be expressed as
\begin{align}
    \begin{split}
        &\frac{d}{dt} \log \rho_\phi(t, x_t) = \frac{\partial}{\partial t} \log \rho_\phi(t, x_t)\\
        &\qquad+ \nabla \log \rho_\phi(t, x_t)\cdot\dot{x}_t + \frac{1}{2}g(t)^2 \nabla^2 \log \rho_\phi(t, x_t),
    \end{split}
\end{align}
where we use the It\^o formula.
Taking into account that $\rho_\phi(t, \cdot)$ is given by the solution of the FPE~\eqref{eq:FPE_forward}, $\mathcal{L}'(t,x_t, \dot{x}_t)$ can be calculated as 
\begin{align}
    &\mathcal{L}_\phi'(t,x_t, \dot{x}_t)\nonumber\\
    &= \frac{1}{2g(t)^2}\|\dot{x}_t - ( u_\phi(t,x_t) - g(t)^2\nabla\log \rho_\phi(t, x_t))\|^2\nonumber\\
    &\qquad- \nabla\cdot ( u_\phi(t,x_t) - g(t)^2\nabla\log \rho_\phi(t, x_t)).
\end{align}
By comparing this form with Eq.~\eqref{eq:appendix_C_path_integral_reverse}, we find that the conditional path probability $\mathbb{P}_\phi[x_{[0,T)}|x_T = z]$ is generated by the reverse-It\^o SDE with drift term $u_\phi(t,x_t) - g(t)^2\nabla\log \rho_\phi(t, x_t)$. That is Eq.~\eqref{eq:SDE_encoder_reverse}. 

\section{Girsanov theorem for reverse-It\^o SDEs and the objective function for the diffusion model}
\label{sec:appendix_objective_DM}
The left-hand side of Eq.~\eqref{eq:Girsanov} can be represented as
\begin{align}
    \begin{split}
        &\mathbb{E}_{\rho_\phi(z)}[D_\text{KL}(\mathbb{P}_\phi[x_{[0,T)}|x_T=z]\|\mathbb{Q}_\theta[x_{[0,T)}|x_T=z])]\\
        &\qquad= \mathbb{E}_{\mathbb{P}_\phi[x_{[0,T]}]}\left[\log\frac{\mathbb{P}_\phi[x_{[0,T)}|x_T=z]}{\mathbb{Q}_\theta[x_{[0,T)}|x_T=z]}\right].
        \label{eq:appendix_D_KL_path}
    \end{split}
\end{align}
By using the descritized represatation of the conditional path probability, Eq.~\eqref{eq:appendix_C_path_probability_reverse_descritized}, we obtain
\begin{align}
    &\log \frac{\mathbb{P}_\phi[x_{[0,T)}|x_T=z]}{\mathbb{Q}_\theta[x_{[0,T)}|x_T=z]}\nonumber\\
    &= \lim_{N\rightarrow \infty}
     -  \sum_{i=0}^{N-1} \frac{\Delta t}{2g(t_{i+1})^2}
    \Bigg[\left\|\frac{x_{i+1} - x_i}{\Delta t} - \tilde{u}_\phi(t_{i+1},x_{i+1})\right\|^2 \nonumber\\
    &\hspace{2.5cm}- \left\|\frac{x_{i+1} - x_i}{\Delta t} - s_\theta(t_{i+1},x_{i+1})\right\|^2\Bigg],
    \label{eq:expand_quotient}
\end{align}
where we denote $\tilde{u}_\phi := u_\phi - g^2\nabla\log \rho_\phi$ for notational simplicity.
The expansion of the square terms gives
\begin{align}
    &\left\|\frac{x_{i+1} - x_i}{\Delta t} - \tilde{u}_\phi(t_{i+1},x_{i+1})\right\|^2\nonumber\\
    &\hspace{2.5cm}- \left\|\frac{x_{i+1} - x_i}{\Delta t} - s_\theta(t_{i+1},x_{i+1})\right\|^2\nonumber\\
    &= -2\frac{x_{i+1} - x_i}{\Delta t} \left\{\tilde{u}_\phi(t_{i+1},x_{i+1})- s_\theta(t_{i+1},x_{i+1})\right\}\nonumber\\
    &\qquad\quad + \{\|\tilde{u}_\phi(t_{i+1},x_{i+1})\|^2 - \|s_\theta(t_{i+1},x_{i+1})\|^2\}
    \label{eq:expand_square}
\end{align}
Since the realizations $x_{[0,T]}$ generated from the reverse SDE~\eqref{eq:SDE_encoder_reverse} only contribute the expectation in Eq.~\eqref{eq:appendix_D_KL_path}, $x_{i+1} - x_i$ can be evaluated as
\begin{align}
    x_{i+1} - x_i =  \tilde{u}_\phi(t_{i+1}, x_{i+1}) \Delta t + g(t_{i+1})\Delta \bar{w}_{i}.
\end{align}
The substitution of the above equation into Eq.~\eqref{eq:expand_square} gives
\begin{align}
    &\left\|\frac{x_{i+1} - x_i}{\Delta t} - \tilde{u}_\phi(t_{i+1},x_{i+1})\right\|^2\nonumber\\
    &\hspace{2.5cm}- \left\|\frac{x_{i+1} - x_i}{\Delta t} - s_\theta(t_{i+1},x_{i+1})\right\|^2\nonumber\\
    &= -2\frac{\Delta \bar{w}_i}{\Delta t}g(t_{i+1})\left\{\tilde{u}_\phi(t_{i+1},x_{i+1})- s_\theta(t_{i+1},x_{i+1})\right\}\nonumber\\
    &\qquad\qquad - \left\|\tilde{u}_\phi(t_{i+1},x_{i+1}) - s_\theta(t_{i+1},x_{i+1})\right\|^2.
\end{align}
Hence, we obtain
\begin{align}
    &\log \frac{\mathbb{P}_\phi[x_{[0,T)}|x_T=z]}{\mathbb{Q}_\theta[x_{[0,T)}|x_T=z]}\nonumber\\
    &= \lim_{N\rightarrow \infty}
    \sum_{i=0}^{N-1} 
    \Bigg[ \frac{\tilde{u}_\phi(t_{i+1},x_{i+1}) - s_\theta(t_{i+1},x_{i+1})}{g(t_{i+1})}\Delta \bar{w}_i\nonumber\\
    &\qquad\qquad + \frac{\left\|\tilde{u}_\phi(t_{i+1},x_{i+1}) -  s_\theta(t_{i+1},x_{i+1})\right\|^2}{2g(t_{i+1})^2}\Delta t\Bigg]\nonumber\\
    &= \int_0^T \frac{\tilde{u}_\phi(t,x_t) - s_\theta(t,x_t)}{g(t)} \rIto d\bar{w}_t \nonumber\\
    &\qquad\qquad+ \frac{1}{2}\int_0^T \frac{\|\tilde{u}_\phi(t,x_t) - s_\theta(t,x_t)\|^2_2}{g(t)^2}\rIto dt,
    \label{eq:Girsanov_reverse}
\end{align}
This is the Girsanov theorem \cite{oksendal2013stochastic} for the reverse-It\^o SDE.
If we take expectation over $\mathbb{P}_\phi[x_{[0,T]}]$ for the both sides of Eq.~\eqref{eq:Girsanov_reverse}, the first term vanishes, and 
we get 
\begin{align}
    &\mathbb{E}_{\mathbb{P}_\phi[x_{[0,T]}]}\left[\log\frac{\mathbb{P}_\phi[x_{[0,T)}|x_T=z]}{\mathbb{Q}_\theta[x_{[0,T)}|x_T=z]}\right]\nonumber\\
    &= \mathbb{E}_{\mathbb{P}_\phi[x_{[0,T]}]} \Bigg[
    \frac{1}{2}\int_0^T \frac{\|\tilde{u}_\phi(t,x_t) - s_\theta(t,x_t)\|^2}{g(t)^2}\rIto dt\Bigg],\nonumber\\
    &= \frac{1}{2}\int_0^T \frac{dt}{g(t)^2}\mathbb{E}_{\rho_\phi(t,x_t)}\left[\|\tilde{u}_\phi(t, x_t) - s_\theta(t,x_t)\|^2\right].
    \label{eq:KL_divergence_conditional_reformed}
\end{align}
Here, the expectation in the last line is taken over the distribution $\rho_\phi(t,x_t)$ at each time step $t$, while the expectation in the second line is over the path probability $\mathbb{P}_\phi[x_{[0,T]}]$. The substitution of Eq.~\eqref{eq:KL_divergence_conditional_reformed} into Eq.~\eqref{eq:appendix_D_KL_path} leads to Eq.~\eqref{eq:Girsanov}.
\section{Reformation of the objective function with implicit score matching}
\label{sec:appendix_reformed_objective}


The first term of the objective function in Eq.~\eqref{eq:objective_function_DM_reformed} can be decomposed as 
\begin{align}
    D_\text{KL}(p_\phi(z)\|\pi(z)) = \mathbb{E}_{p_\phi(z)}\left[\log p_\phi(z) \right] - \mathbb{E}_{p_\phi(z)}\left[ \log \pi(z)\right],
\end{align}
where $p_\phi(z)$ is the marginal distribution of $\mathbb{P}_\phi[x_{[0,T]}]$ at $t = T$. 
The first term can be rewritten as
\begin{align}
    &\mathbb{E}_{p_\phi(z)}\left[\log p_\phi(z) \right]\nonumber\\
    &= \mathbb{E}_{\mu(x)}[\log \mu(x)] + \int_0^T dt\, \frac{d}{dt}\mathbb{E}_{\rho_\phi(t, x)}\left[ \log \rho_\phi(t, x) \right].
    \label{eq:app_entropy_p-phi}
\end{align}
By calculating the integrand in the second term in the last line, we have
\begin{align}
    \frac{d}{dt}\mathbb{E}&_{\rho_\phi(t, x)}\left[ \log \rho_\phi(t, x) \right]\nonumber\\
    &= \frac{d}{d t} \int dx\, \rho_\phi(t,x)\log \rho_\phi(t,x)\nonumber\\
    &= \int dx \frac{\partial \rho_\phi(t,x)}{\partial t}\log \rho_\phi(t,x) + \frac{d}{d t}\int dx \rho_\phi(t,x)
    \label{eq:app_e_entropy_p-phi}
\end{align}
The second term of the RHS vanishes because the probability distribution $\rho_\phi(t, x)$ is normalized for any $t$. Moreover, since $\rho_\phi$ is the solution of the encode FPE \eqref{eq:FPE_forward}, the substitution of the FPE~\eqref{eq:FPE_forward}
into the first term of Eq.~\eqref{eq:app_e_entropy_p-phi} gives
\begin{align}
    &\frac{d}{dt}\mathbb{E}_{\rho_\phi(t, x)}\left[ \log \rho_\phi(t, x) \right]\nonumber\\
    &= -\int dx\, \nabla\cdot\left[\left(u_\phi - \frac{1}{2}g^2\nabla\log \rho_\phi\right)\rho_\phi\right]\log \rho_\phi\nonumber\\
    &= \int dx\, \rho_\phi\left(u_\phi - \frac{1}{2}g^2\nabla\log \rho_\phi\right)\cdot\nabla\log \rho_\phi\nonumber\\
    &= \frac{1}{2g^2}\,\mathbb{E}_{\rho_\phi}\left[2u_\phi\cdot (g^2\nabla \log \rho_\phi) - \|g^2\nabla\log \rho_\phi\|^2\right].
    \label{eq:app_integrand_entropy_p-phi}
\end{align}
Here, we used the integration by parts and ignored the surface term to get the third line.
Therefore, we obtain
\begin{align}
    &D_\text{KL}(p_\phi\|\pi) = \mathbb{E}_{\mu}[\log \mu] - \mathbb{E}_{p_\phi}[ \log \pi]\nonumber\\
    &+ \frac{1}{2}\int_0^T \frac{dt}{g^2}\mathbb{E}_{\rho_\phi}[2u_\phi\cdot(g^2\nabla\log \rho_\phi) - \|g^2 \nabla\log \rho_\phi\|^2].
    \label{eq:app_expansion_first_term}
\end{align}

The second term of the objective function in Eq.~\eqref{eq:objective_function_DM_reformed} can be expanded as 
\begin{align}
    &\frac{1}{2}\int_0^T \frac{dt}{g^2}\mathbb{E}_{\rho_\phi}\left[\|u_\phi-g^2\nabla \log \rho_\phi-s_\theta\|^2\right]\nonumber\\
    &=\frac{1}{2}\int_0^T \frac{dt}{g^2}\mathbb{E}_{\rho_\phi}\big[\|u_\phi-s_\theta\|^2 \nonumber\\
    &\quad + \|g^2\nabla \log \rho_\phi\|^2 - 2 (u_\phi-s_\theta)\cdot(g^2\nabla \log \rho_\phi)\big].
    \label{eq:app_expansion_second_term}
\end{align}
By combining Eqs.~\eqref{eq:app_expansion_first_term} and~\eqref{eq:app_expansion_second_term}, the objective function in Eq.~\eqref{eq:objective_function_DM_reformed} can be reformed as
\begin{align}
    &D_\text{KL}(p_\phi\|\pi)\nonumber\\
    &+\frac{1}{2}\int_0^T \frac{dt}{g(t)^2}\mathbb{E}_{\rho_\phi(t,x_t)}\left[\|\tilde{u}_\phi(t, x_t) - s_\theta(t,x_t)\|^2\right]\nonumber\\
    &\mathbb{E}_{\mu}[\log \mu] - \mathbb{E}_{p_\phi}[ \log \pi]\nonumber\\
    &+ \frac{1}{2}\int_0^T \frac{dt}{g^2}\mathbb{E}_{\rho_\phi}\big[\|u_\phi-s_\theta\|^2 + 2 s_\theta\cdot(g^2\nabla \log \rho_\phi)\big].
\end{align}
Because the first term $\mathbb{E}_{\mu}[\log \mu]$ does not contain the parameters $(\phi, \theta)$, we can ignore it in the optimization problem. Furthermore, for any probability distribution $p(x)$ with the condition $\lim_{\|x\|\to \infty} p(x) = 0$ and test function $f(x)$, the following equality holds:
\begin{align}
    \mathbb{E}_{p(x)}[\nabla \log p(x)\cdot f(x)]
    = -\mathbb{E}_{p(x)}[\nabla\cdot f(x)].
\end{align}
This leads to Eq.~\eqref{eq:objective_function_DM_computation} as
\begin{align}
    &D_\text{KL}(p_\phi\|\pi)\nonumber\\
    &+\frac{1}{2}\int_0^T \frac{dt}{g(t)^2}\mathbb{E}_{\rho_\phi(t,x_t)}\left[\|\tilde{u}_\phi(t, x_t) - s_\theta(t,x_t)\|^2\right]\nonumber\\
    &= - \mathbb{E}_{p_\phi}[\log \pi] + \frac{1}{2}\int_0^T \frac{dt}{g^2}\mathbb{E}_{\rho_\phi}\big[\|u_\phi-s_\theta\|^2 - 2 g^2 \nabla\cdot s_\theta\big].
\end{align}
The reformulation from the original objective function in Eq.~\eqref{eq:objective_function_DM_reformed} to Eq.~\eqref{eq:objective_function_DM_computation} called implicit score matching \cite{hyvarinen2005estimation}. 

\section{Probability-flow ODE}
\label{sec:appendix_probability-flow_ODE}

A FPE corresponding to the decode SDE \eqref{eq:SDE_decoder} is given by
\begin{align}
    &\frac{\partial \tilde{\rho}(t, x)}{\partial t}\nonumber\\ 
    &= -\nabla\cdot\left[ \left( s_\theta(t,x) \right)\tilde{\rho}(t,x) \right] - \frac{1}{2}g(t)^2\nabla^2\tilde{\rho}(t,x)\nonumber\\
    &= -\nabla\cdot\left[\left\{  s_\theta(t, x)+ \frac{1}{2}g(t)^2\nabla\log\tilde{\rho}(t,x) \right\}\tilde{\rho}(t,x) \right].
    \label{eq:FPE_decoder}
\end{align}
By forcussing the last line, we notice that Eq.~\eqref{eq:FPE_decoder} is also derived from the ODE with the vector field $s_\theta + 2^{-1}g^2\nabla\log \tilde{\rho}$:
\begin{align}
    \frac{dX_t}{dt} = s_\theta(t, X_t)+ \frac{1}{2}g(t)^2\nabla\log\tilde{\rho}_\theta(t,X_t).
    \label{eq:app_probability-flow_ODE}
\end{align}
Here, $\tilde{\rho}_\theta(t,x)$ is the solution of the FPE \eqref{eq:FPE_decoder} with the terminal condition $\tilde{\rho}_\theta(T,\cdot) = \pi(\cdot)$. That is Eq. \eqref{eq:probability-flow_ODE_1}.

If the optimization in Eq.~\eqref{eq:objective_function_DM_reformed} is sufficiently achieved, 
the path probability in the decoding process can be approximated by the one of the encoding process: $\mathbb{Q}_{\theta^*}[x_{[0,T]}]\simeq \mathbb{P}_{\phi^*}[x_{[0,T]}]$. Therefore, for the marginal distribution, $\tilde{\rho}_{\theta^*} \simeq \rho_{\phi^*}$ holds. 
Furthermore, owing to the drift matching in Eq.~\eqref{eq:objective_function_DM_reformed}, we obtain
 $g^2\nabla \log \rho_{\phi^*} \simeq s_{\theta^*} - u_{\phi^*}$.
Thus, under the optimal condition in Eq.~\eqref{eq:objective_function_DM_reformed}, the above ODE~\eqref{eq:app_probability-flow_ODE} can be rewritten as
\begin{align}
    \frac{dX_t}{dt} 
    &\simeq u_{\phi^*}(t,X_t) + \frac{1}{2}\nabla \log \rho_{\phi^*}(t, X_t)\nonumber\\
    &\simeq u_{\phi^*}(t,X_t) + \frac{1}{2}[u_{\phi^*}(t,X_t)-s_{\theta^*}(t, X_t)]\nonumber\\
    &= \frac{1}{2}[u_{\phi^*}(t,X_t)+s_{\theta^*}(t, X_t)],
\end{align}
which is called the probability-flow ODE~\cite{chen2022likelihood}.

\bibliography{99_references}

\end{document}